\documentclass[sigconf, screen, authorversion]{acmart}
\renewcommand
\footnotetextcopyrightpermission[1]{}
\DeclareUnicodeCharacter{2206}{$\Delta$}
\AtBeginDocument{%
  }
    
\setcopyright{none}

\usepackage[utf8]{inputenc}
\usepackage{algorithm}
\usepackage{algorithmic}
\usepackage{graphicx}
\usepackage{subcaption}
\usepackage{hyperref}

\begin{document}

\title{HIRO: Hierarchical Information Retrieval Optimization}

\author{Krish Goel}
\affiliation{%
  \city{New Delhi}
  \country{India}}
\email{krishgoel3@gmail.com}

\author{Mahek Chandak}
\affiliation{%
  \city{Kolkata}
  \country{India}
}
\email{mahekchandak99@gmail.com}

\renewcommand{\shortauthors}{Krish et al.}

\begin{abstract}
Retrieval-Augmented Generation (RAG) has revolutionized natural language processing by dynamically integrating external knowledge into Large Language Models (LLMs), addressing their limitation of static training datasets. Recent implementations of RAG leverage hierarchical data structures, which organize documents at various levels of summarization and information density. This complexity, however, can cause LLMs to "choke" on information overload, necessitating more sophisticated querying mechanisms. In this context, we introduce Hierarchical Information Retrieval Optimization (HIRO), a novel querying approach that employs a Depth-First Search (DFS)-based recursive similarity score calculation and branch pruning. This method uniquely minimizes the context delivered to the LLM without informational loss, effectively managing the challenge of excessive data. HIRO's refined approach is validated by a 10.85\% improvement in performance on the NarrativeQA dataset.
\end{abstract}



\maketitle
\pagestyle{plain}

\section{Introduction}
The advent of Large Language Models (LLMs) has brought about significant transformation in artificial intelligence and natural language processing, enabling advancements in tasks such as text generation, translation, and question-answering \cite{bubeck2023sparks}. As these models increase in size and complexity, they have evolved into highly effective standalone knowledge repositories, embedding vast amounts of factual information within their parameters \cite{petroni2019language, talmor2020olympics}. However, LLMs are inherently constrained by the static nature of their training datasets, which limits their adaptability to continuously evolving real-world information, where the breadth and depth of knowledge required far exceed any static training corpus \cite{minaee2024large}. This limitation underscores a critical gap in LLM design, where fine-tuning falls short, particularly when updates and domain-specific information are necessary \cite{lewis2020retrieval}.

In response to these challenges, the field has witnessed the emergence of Retrieval-Augmented Generation (RAG) models, evolving into what are now known as Retrieval-Augmented Language Models (RALMs). These new paradigms enhance LLMs by integrating dynamic external knowledge through sophisticated retrieval mechanisms, effectively addressing the inherent limitations of static knowledge bases \cite{lewis2020retrieval}. Inspired by the success of open-domain question answering systems, RALMs utilize indexed text databases to enhance model responses by presenting retrieved information alongside input queries \cite{ram2023context, akyurek2022towards}. This integration not only boosts the models’ ability to provide current, domain-specific knowledge but also enhances interpretability and source traceability, offering a stark contrast to the opaque parametric knowledge of traditional LLMs.

Amidst this evolving landscape, novel approaches to information retrieval, such as the use of hierarchical data structures for organizing documents, represent significant advancements. Techniques like RAPTOR: Recursive Abstractive Processing for Tree-Organized Retrieval, which clusters and summarizes texts into a tree structure, demonstrate the potential to overcome traditional limitations by providing both granular and high-level insights \cite{kanagavalli2016study, wang2024detect, he2024retriever, sarthi2024raptor}. However, the adoption of hierarchical data structures poses challenges, particularly the overwhelming amount of context they can return, which may not always align with the query requirements \cite{sarthi2024raptor}. For queries demanding extensive information, this can result in insufficient data being returned, whereas for simpler inquiries, it might overwhelm LLMs with excessive context. This static definition of 'k' in conventional querying does not account for the variability in information needs, often leading to an information "choke" in LLMs \cite{liu2024lost, sun2021long, levy2024same}.

To address this, our research introduces a novel querying technique specifically designed to optimize the use of hierarchical data structures within retrieval-augmented generation systems. By implementing a depth-first search (DFS)-based recursive similarity thresholding and branch pruning, our approach aims to selectively minimize the context delivered to LLMs. This strategy intends to mitigate the risk of information overload, ensuring that LLMs receive a distilled yet comprehensive information set, thereby enabling more accurate and efficient response generation.
\begin{figure*}
  \centering
  \includegraphics[width=\linewidth]{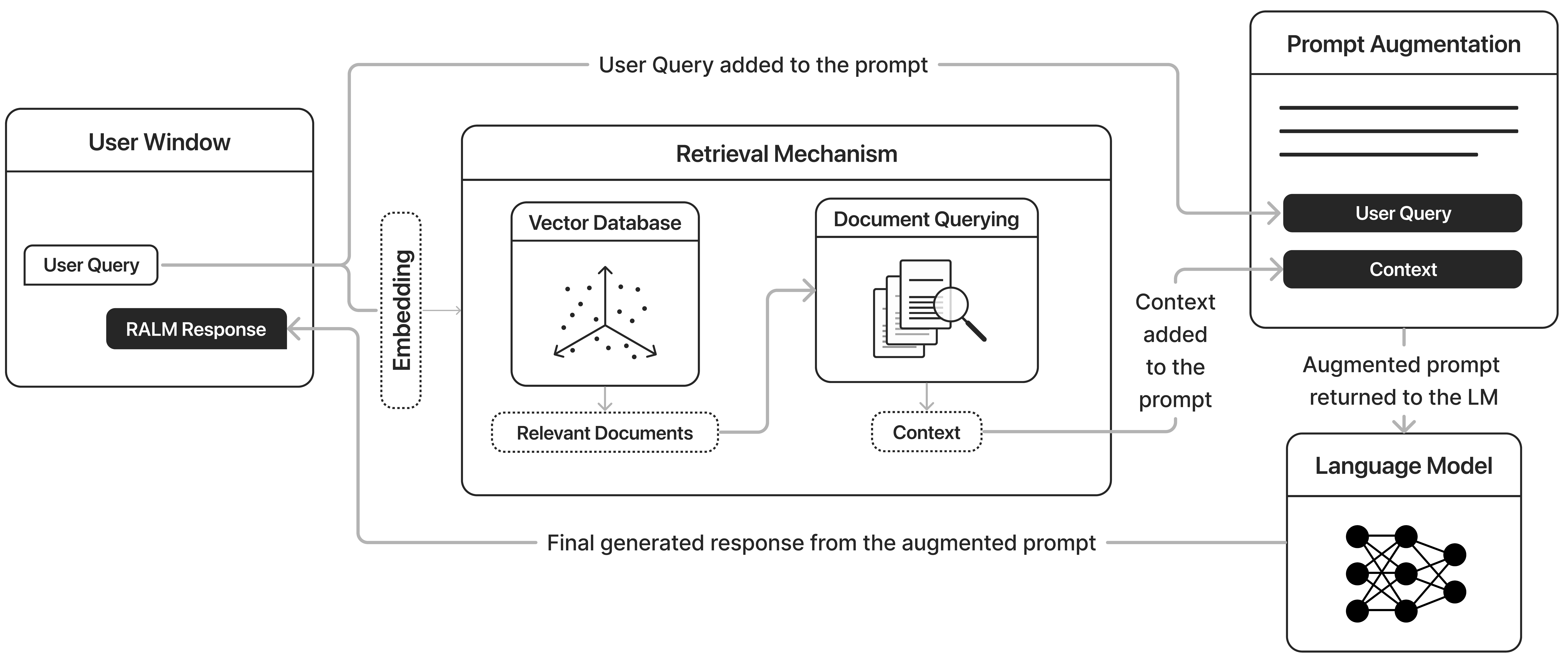}
  \caption{Architectural Overview of a Retrieval-Augmented Language Model (RALM). A query is inputted, processed through an embedding model to generate a vector representation, which is then matched in a vector database to retrieve relevant contexts. These contexts are fed, along with the original query, into the LLM, resulting in an informed response.}
  \Description{A query is inputted, processed through an embedding model to generate a vector representation, which is then matched in a vector database to retrieve relevant contexts. These contexts are fed, along with the original query, into the LLM, resulting in an informed response.}
\end{figure*}

\section{Related Work}
\textbf{Retrieval-Augmented Generation (RAG)} \cite{lewis2020retrieval}, enhances Large Language Models (LLMs) by integrating real-world data from external databases, thus establishing Retrieval Augmented Language Models (RALMs). This approach ensures generated content is both contextually relevant and factually accurate, addressing the issue of LLMs producing plausible but incorrect ”hallucinated” in- formation \cite{shuster2021retrieval}. By enhancing the accuracy and reliability of AI-generated text, RAG heralds a new era of trustworthy machine-generated communications. It's widespread use in information retrieval and question-answering showcases its utility. However, challenges remain, particularly in scalability and efficiency when dealing with large datasets and complex structures. These issues highlight the need for ongoing research in optimization techniques, an area where Hierarchical Information Retrieval shows potential.

\textbf{The adoption of hierarchical structures in RAG models}, exemplified by RAPTOR \cite{sarthi2024raptor}, represents a pivotal shift towards more organized and efficient information storage and retrieval. RAPTOR’s approach, through recursive summarisation to create a hierarchical tree, facilitates nuanced access to information at various levels of abstraction. This innovation not only improves coherence and information density tailored to specific tasks but also encourages exploration into other hierarchical frameworks like graphs for enhanced document interrelation preservation. These advancements signal a new phase in computational linguistics, focusing on sophisticated, hierarchical data use for better knowledge representation and retrieval. However, existing hierarchical models often face challenges in balancing information density and retrieval efficiency, leading to potential information overload or loss.

\textbf{Dynamic Querying Mechanisms} in RAG Models address the evolving nature of queries where traditional static retrieval methods, such as TF-IDF and BM25, have been complemented by more adaptive techniques. The Tree Traversal method progressively selects the top-k most relevant nodes from each layer of the hierarchical data based on similarity, allowing for adjustments in depth and number of nodes to control the information’s specificity and breadth. Conversely, the Collapsed Tree approach flattens the hierarchy to allow for simultaneous comparison of all nodes. \hyperref[fig:treetraversal]{Figure 5} and \hyperref[fig:collapsedtree]{Figure 6} illustrate the workings of these algorithms, providing visual representations.

Despite these innovations, a limitation arises from the static quantity of data retrieved, which may not always align with the query’s needs, leading to potential information overload for LLMs. This static k value, irrespective of query complexity, can hamper LLMs’ performance by either providing insufficient data for complex queries or overwhelming them with excessive context for simpler ones. Additionally, returning both parent and child nodes due to similar matching levels often results in inefficient, redundant information. Optimizing the retrieval process to avoid such duplication is crucial for improving LLM performance in hierarchical document structures.

Our proposed approach, HIRO (Hierarchical Information Retrieval Optimization), addresses these shortcomings by employing Recursive Similarity Thresholding and Branch Pruning for dynamic query response optimization. Leveraging the Similarity Threshold (\(S\)) and Delta Threshold (\(\Delta\)) hyperparameters, HIRO customizes the retrieval process to match query complexities. The Similarity Threshold (\(S\)) filters document graphs based on similarity scores, while the Delta Threshold (\(\Delta\)) refines relevance by pruning branches. This flexible approach allows for the fine-tuning of hyperparameters at the document or corpus level to better meet specific informational needs and contexts.

\begin{figure*}
  \centering
  \includegraphics[width=0.9\linewidth]{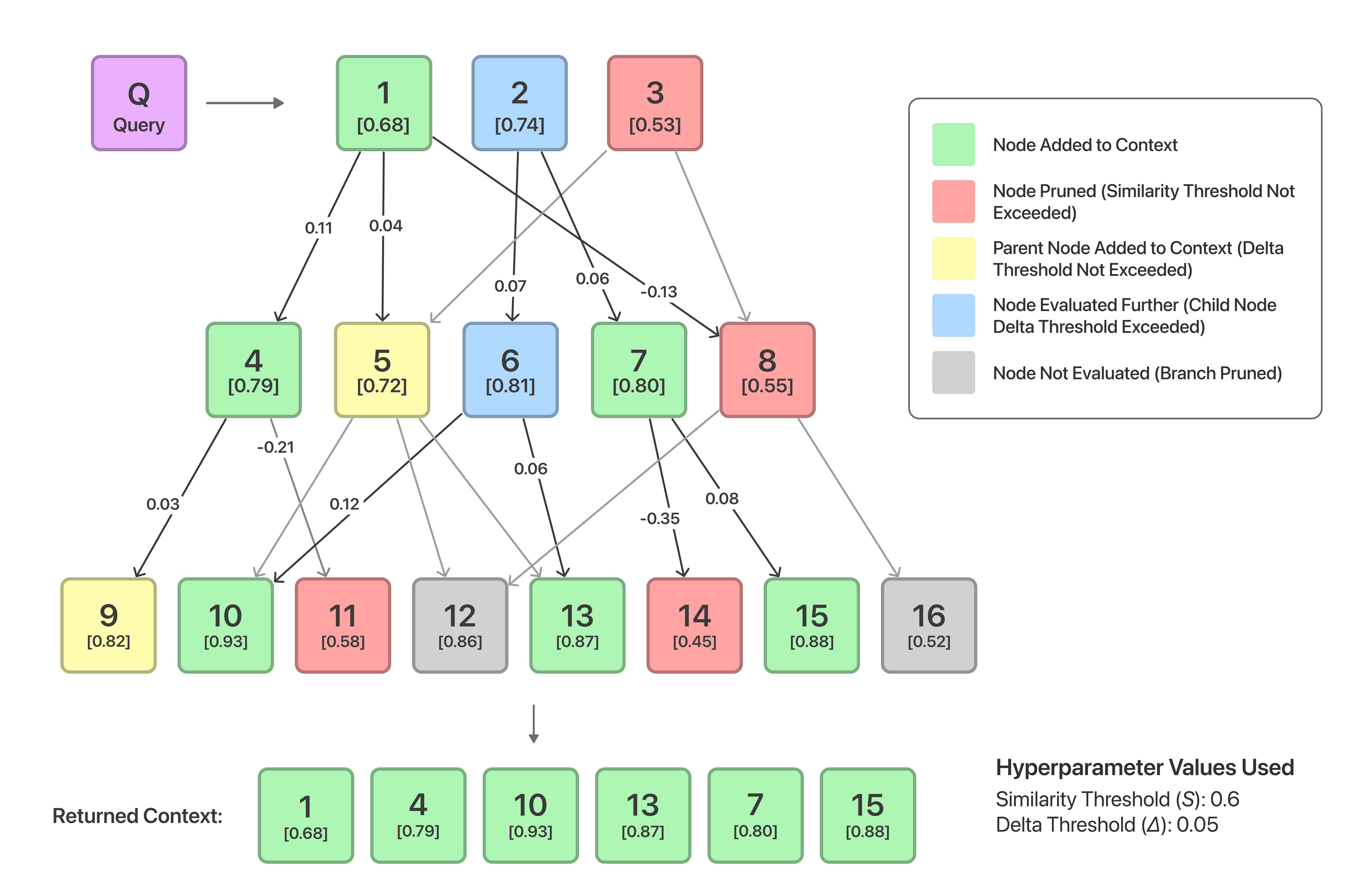}
  \caption{HIRO Querying Mechanism. HIRO's recursive process employs Selection and Delta Thresholds to filter and prune document graphs for optimized information retrieval. It highlights how relevant contexts are dynamically identified and retrieved based on a balance of specificity and information breadth, tailored to each unique query.}
  \Description{HIRO's recursive process employs Selection and Delta Thresholds to filter and prune document graphs for optimized information retrieval. It highlights how relevant contexts are dynamically identified and retrieved based on a balance of specificity and information breadth, tailored to each unique query.}
\end{figure*}

\section{Methodology}

Our approach introduces Recursive Similarity Score Calculation and Branch Pruning to optimize query responses through dynamic retrieval processes customised to the specifics of each query. This method is grounded in the utilization of two hyperparameters: the Selection Threshold ($S$) and the Delta Threshold (\(\Delta\)), which combine to create a dynamic retrieval process depending on the complexity of the query.

The detailed pseudocode for our HIRO approach is outlined in \hyperref[alg:hiroquerying]{Algorithm 1}. For a version optimized for deployment in production environments, please refer to \hyperref[alg:alternatehiroquerying]{Algorithm 3} in \hyperref[sec:alternatepseudocode]{Appendix B}, which includes both the primary HIRO algorithm and the auxiliary \hyperref[alg:evaluatechildren]{Algorithm 4} for evaluating child nodes.

\subsection{Recursive Similarity Score Calculation}

The Selection Threshold (S) serves as an initial filter, identifying document graphs for further exploration based on similarity scores between the query and parent nodes. Only graphs with parent nodes surpassing this threshold proceed, ensuring focus on potentially relevant content.

The process begins at the root layer of the hierarchical data structure, where the similarity between the query embedding \( Q \) and the embeddings of all nodes \( N_i \) present at this initial layer is computed. The similarity score \( S(Q, N_i) \) is determined using cosine similarity, which measures the cosine of the angle between two non-zero vectors, providing a value between -1 and 1:

\[ S(Q, N_i) = \frac{Q \cdot N_i}{\|Q\| \|N_i\|} \]

Nodes whose similarity scores exceed the predefined Selection Threshold \( S \) are earmarked for further exploration. This step ensures that only nodes with a significant level of relevance to the query are considered in subsequent stages.

\[ \text{Earmarked Nodes} = \{ N_i \mid S(Q, N_i) > S \} \]

\subsection{Branch Pruning}

The Delta Threshold (\(\Delta\)) refines the search by evaluating each child node within accepted graphs. This step is crucial in eliminating less relevant data, thus enhancing the efficiency and relevance of the retrieval process.

For each earmarked parent node \( N_i \), the similarity between the query embedding \( Q \) and each child node \( N_j \) is calculated. The child nodes' similarity scores \( S(Q, N_j) \) are compared to their respective parent nodes' scores. The difference in similarity, denoted as \( \Delta S \), is computed as follows:

\[ \Delta S = S(Q, N_j) - S(Q, N_i) \]

A child node is marked for further recursive evaluation if \( \Delta S \) exceeds the Delta Threshold \( \Delta \):

\[ \Delta S > \Delta \]

If a child node does not meet this criterion, the parent node is retained for context. This selective pruning ensures that only the most promising branches are explored, enhancing the efficiency of the retrieval process.

\subsection{Dynamic Retrieval Process}

The dynamic retrieval process aggregates the selected nodes to form an optimized and coherent context for the language model (LLM). This approach balances the retention of pertinent information with the need to avoid overwhelming the system with excessive detail.

\subsubsection{Termination Criteria}

The recursive search terminates under two conditions: when the current node is a leaf node, or when no further nodes meet the threshold criteria. Specifically, the search ends if the current node does not have any children or if the similarity score improvement does not exceed the Delta Threshold.

\subsubsection{Context Aggregation}

Upon termination of the search, the selected nodes are aggregated to form the final context for the LLM. Nodes \( N_k \) are included if their similarity \( S(Q, N_k) \) to the query exceeds the Selection Threshold \( S \) and if each child node \( N_j \) shows a similarity gain \( \Delta S_{kj} = S(Q, N_j) - S(Q, N_k) \) above the Delta Threshold \( \Delta \). This ensures the LLM receives a coherent and comprehensive set of relevant information.

\[
\text{Final Context} = \left\{ N_k \mid S(Q, N_k) > S \text{ and } \forall N_j, \, \Delta S_{kj} > \Delta \right\}
\]

\subsection{Hyperparameter Impact}

The Selection Threshold \( S \) governs the scope of data examined, controlling the breadth of information explored. In contrast, the Delta Threshold \( \Delta \) governs the extent of information examined, managing the depth of the search.

While traversing deeper into the tree, similarity scores naturally escalate. Nonetheless, returning excessively granular context becomes impractical due to the resultant escalation in context length and computational demands. The Delta Threshold serves as the pivotal point that delineates the balance between retaining pertinent information and mitigating the risk of inundating the system with excessive detail.

The interplay between \( S \) and \( \Delta \) is critical:

\begin{itemize}
    \item \textbf{Selection Threshold \( S \)}: A higher \( S \) means only highly similar nodes are considered, narrowing the initial search but potentially missing relevant sub-nodes.
    \item \textbf{Delta Threshold \( \Delta \)}: A higher \( \Delta \) means only significant improvements in similarity are considered, reducing the depth of search and focusing on more promising nodes.
\end{itemize}

By fine-tuning these hyperparameters, the retrieval process can be dynamically adapted to the complexity and requirements of each specific query, ensuring efficient and relevant information extraction.

\begin{algorithm}
\caption{HIRO Querying}
\label{alg:hiroquerying}
\begin{algorithmic}[0]
\STATE\textbf{Input:} Query $query$, Tree $tree$, Selection Threshold $S$, Delta Threshold $\Delta$
\STATE\textbf{Output:} Context Nodes
\end{algorithmic}
\begin{algorithmic}[1]
\STATE $context \leftarrow \text{[]}$
\STATE $nodes\_to\_evaluate \leftarrow tree\text{.layer[0]}$
\WHILE{$nodes\_to\_evaluate$ is not empty}
    \STATE $node \leftarrow nodes\_to\_evaluate\text{.pop()}$
    \IF{$node$.parent exists}
        \STATE $parent\_score \leftarrow \text{similarity}(query, node.parent)$
    \ELSE
        \STATE $parent\_score \leftarrow \text{0}$
    \ENDIF
    \STATE $score \leftarrow \text{similarity}(query, node)$
    \STATE $delta \leftarrow score - parent\_score$
    \IF{(($score$ > $S$ \text{and} $node$\text{.is\_leaf()}) \text{or} ($delta$ > $\Delta$))}
        \STATE $context$.append($node$.content)
    \ELSE
        \STATE $nodes\_to\_evaluate\text{.extend(}node\text{.children)}$
    \ENDIF
\ENDWHILE
\STATE \textbf{return} $context$
\end{algorithmic}
\end{algorithm}

\section{Experiments}
\subsection{Datasets}
We assessed the performance of HIRO paired with RAPTOR using two prominent question-answering datasets: NarrativeQA and QuALITY.

NarrativeQA is an extensive English-language dataset that includes 1,567 stories, drawn from both books and movie scripts. The dataset is accompanied by 46,765 questions, which are crafted based on human-generated abstractive summaries of the stories. The stories range between 50,000 and 100,000 words \cite{kocisky2018narrativeqa}. This dataset is designed to evaluate a system’s ability to focus on key narrative elements while filtering out irrelevant details. It tests systems’ ability to prune irrelevant details for key story elements, showcasing querying optimization for narrative understanding and information retrieval. Evaluation metrics include ROUGE, which measures recall of n-grams; BLEU-1 and BLEU-4, which assess the precision of n-grams in generated summaries; and METEOR, which evaluates precision and recall by considering synonyms and stemming.

The QuALITY dataset (Question Answering with Long Input Text, Yes!) is designed for evaluating multiple-choice question answering tasks on lengthy English texts. It contains articles with lengths ranging from 2,000 to 8,000 tokens, providing a diverse range of textual complexity \cite{pang2021quality}. The dataset includes 6,737 questions, with a challenging subset named QuALITY-HARD, which features 3,360 particularly difficult questions. This subset is intended to assess the capability of QA models to extract and utilize relevant information from extensive text passages effectively. Performance is measured using the F1 score, which balances precision and recall, and Accuracy, which determines the proportion of correct answers. These metrics help evaluate how well different pruning methods and information retrieval strategies perform in identifying crucial details within long documents.

\subsection{Controlled Baseline Comparisons}

To establish baseline performance, we utilize RAPTOR with SBERT \cite{reimers2019sentence} as the embedding model and GPT-3.5 Turbo \cite{brown2020language} as the reader/language model. This configuration represents the state-of-the-art approach in retrieval-augmented generation using RAPTOR. We systematically compare the performance of RAPTOR combined with SBERT and GPT-3.5 Turbo, both with and without the implementation of HIRO querying.

\textbf{Baseline Configuration:} Performance is evaluated using RAPTOR with SBERT and GPT-3.5 Turbo. RAPTOR traditionally employs a method known as Collapsed Tree Querying, where it flattens the hierarchical tree structure and selects the top-$k$ nodes based on similarity scores. This conventional approach is compared against the HIRO-enhanced method, which incorporates hierarchical querying to optimize the retrieval process.

\textbf{Comparison with HIRO Querying:} We analyze the performance difference between the conventional Collapsed Tree Querying approach and the HIRO querying method. The primary focus is on understanding how HIRO's dynamic traversal and adaptive thresholds improve retrieval efficiency and information relevance compared to the static approach of selecting top-$k$ nodes.

\textbf{Token Length and Computational Trade-offs:} We also examine the token length of the final context returned to the language model (LM) after querying. This analysis provides insights into the trade-off between computational resources and performance gains. By evaluating how the length of the context affects performance and computational load, we aim to understand the efficiency of different querying methods in managing token limits and processing requirements.

\textbf{Non-linear Performance Degradation:} Research indicates that the performance of large language models (LLMs) degrades non-linearly with increasing prompt token length \cite{levy2024same}. To account for this, we present normalized performance scores for both conventional and HIRO querying methods. These scores are adjusted by the logarithm of the context length (in tokens) to offer a clearer picture of efficiency relative to context size. This normalization helps in comparing the effectiveness of querying methods while considering the impact of prompt length on model performance.

\textbf{Computational Time Complexity:} Finally, we present the time complexity analysis for the conventional Tree Querying Methods and HIRO. This analysis highlights the computational cost associated with each approach, providing a detailed comparison of their efficiency.

\subsection{Experimental Setup}
For each dataset, we fine-tune the values of the two hyperparameters - Selection Threshold ($S$) and Delta Threshold ($\Delta$), to optimize HIRO's performance. Bayesian Optimization is utilized for fine-tuning, with insights provided in \hyperref[sec:bayesianoptimisation]{Appendix C}.

\subsection{Experimental Results}
\textbf{Model Performance on Sequence Generation Tasks:} \hyperref[tab:contextlength]{Table 1} shows the average context length in tokens returned by different querying mechanisms with the RAPTOR model on the NarrativeQA dataset. The HIRO querying mechanism strikes an optimal balance by returning shorter contexts than tree traversal but longer than collapsed tree querying, suggesting an effective trade-off between information overload and sufficiency.

\begin{table}
    \caption{Comparison of Average Context Length (in tokens) returned by Tree Traversal, HIRO and Collapsed Tree Querying on the NarrativeQA Dataset.}
    \label{tab:contextlength}
    \centering
    \begin{tabular}{c c} \toprule
        \textbf{Model Configuration} &
        \textbf{Context Length} \\
        \midrule
        RAPTOR with Tree Traversal Querying &
        2762.46115288\\
        RAPTOR with HIRO Querying &
        2253.345865\\
        RAPTOR with Collapsed Tree Querying &
        1864.293445\\
        \bottomrule
    \end{tabular}
\end{table}

\begin{table}
    \caption{Comparison of Time Complexities of Hierarchical Querying Algorithms}
    \label{tab:timecomplexity}
    \centering
    \begin{tabular}{c c}
        \toprule
        \textbf{Algorithm} &
        \textbf{Time Complexity} \\
        \midrule
        HIRO & \(O(n)\)\\
        Tree Traversal & \(O(n \log m)\)\\
        Collapsed Tree & \(O(n \log n)\)\\
        \bottomrule
    \end{tabular}
\end{table}

\begin{table*}
    \caption{NarrativeQA Dataset: Absolute Performance Metrics and Efficiency-Adjusted Scores. Efficiency scores are adjusted using the logarithm of context length (in tokens) to account for loss in performance with increasing context length.}
    \label{tab:narrativeqa}

    \centering
    \begin{tabular}{c c c c c c} \toprule
        \textbf{Model} & \textbf{Score Type} & \textbf{ROUGE-L F1} & \textbf{BLEU-1} & \textbf{BLEU-4} & \textbf{METEOR} \\
        \midrule
        RAPTOR with Collapsed Tree Querying & Absolute Performance & 0.096542025 & 0.165350602 & 0.084131409 & 0.125686168 \\
        RAPTOR with HIRO Querying & Absolute Performance & 0.120835776 & 0.173288844 & 0.089863157 & 0.089863157 \\
        \midrule
        RAPTOR with Collapsed Tree Querying & Efficiency Adjusted & 0.012819892	& 0.021957038 & 0.011171877 & 0.016689966\\
        RAPTOR with HIRO Querying & Efficiency Adjusted & 0.015651944 & 0.022446228 & 0.011640039 & 0.017361359\\
        \bottomrule
    \end{tabular}
\end{table*}

\begin{table*}
    \caption{Absolute Performance Metrics of Collapsed Tree and HIRO Querying on the QuALITY Dataset.}
    \label{tab:quality}
    \centering
    \begin{tabular}{c c c c c} \toprule
        \textbf{Model} & \textbf{Accuracy} & \textbf{Precision} & \textbf{Recall} & \textbf{F1 Score} \\
        \midrule
        RAPTOR with Collapsed Tree Querying & 0.186201 & 0.476749 & 0.488198 & 0.480011\\
        RAPTOR with HIRO Querying & 0.182493 & 0.449404 & 0.444282 & 0.44546\\
        \bottomrule
    \end{tabular}
\end{table*}

\textbf{Model Performance on NarrativeQA Dataset:} \hyperref[tab:narrativeqa]{Table 2} presents absolute performance and efficiency adjusted metrics comparison of RAPTOR with collapsed tree querying vs HIRO querying on the NarrativeQA dataset using SBERT and GPT-3.5 Turbo. HIRO querying outperformed conventional querying across all metrics.

\textbf{Model Performance on the QuALITY Dataset:} \hyperref[tab:quality]{Table 3} compares the absolute accuracy, precision, recall and F1 score of RAPTOR with HIRO querying against Collapsed Tree querying on the QuALITY dataset. HIRO querying showed slightly lower accuracy and F1 score, compared to collapsed tree querying.


\begin{figure*}[!h]
\centering
  \begin{subfigure}[c]{0.49\textwidth}
    \includegraphics[width=0.8\textwidth]{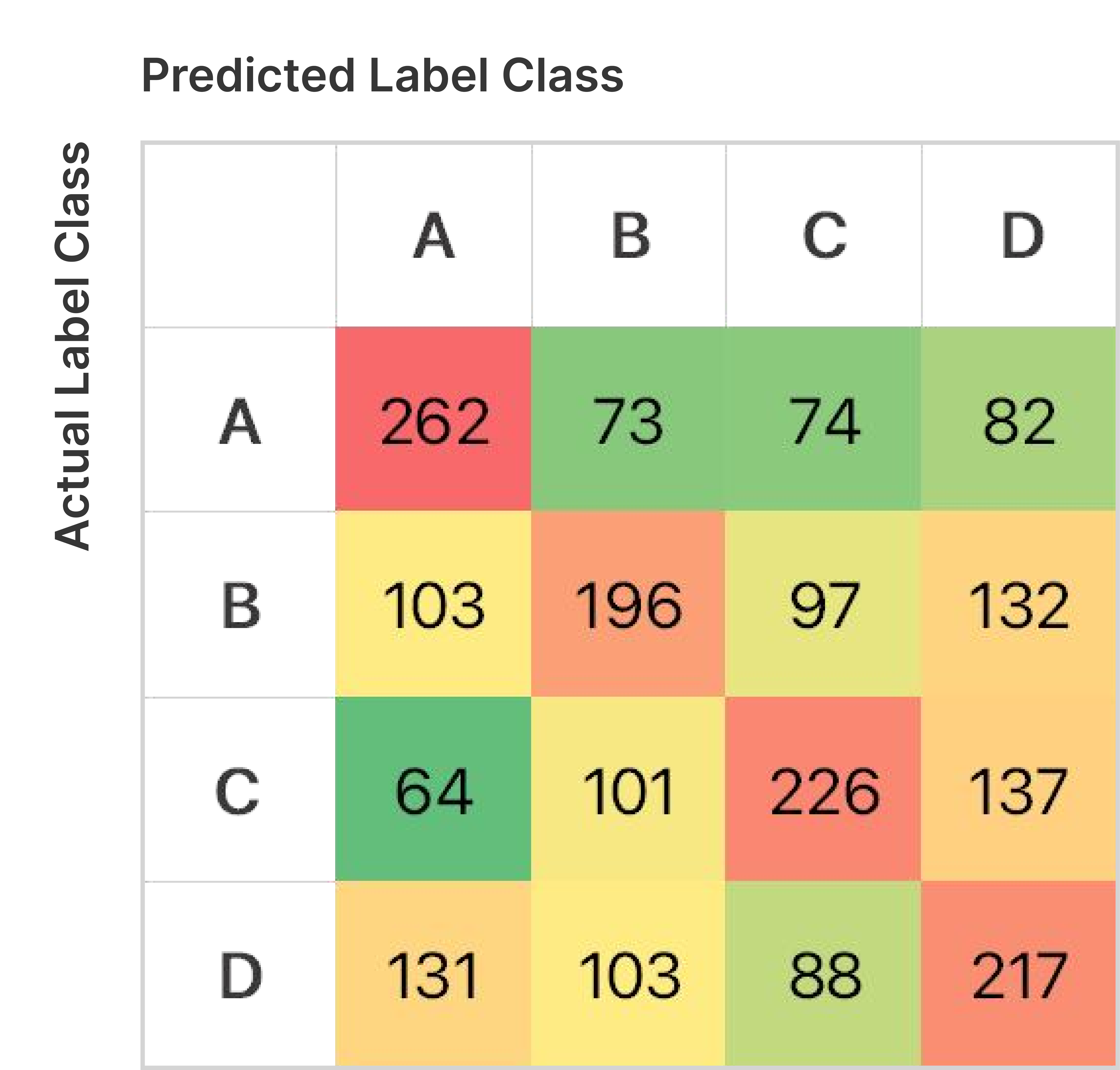}
    \caption{\textbf{RAPTOR with HIRO Querying}}
    \label{fig:confusionmatrixhiro}
  \end{subfigure}
  \begin{subfigure}[c]{0.49\textwidth}
    \includegraphics[width=0.8\textwidth]{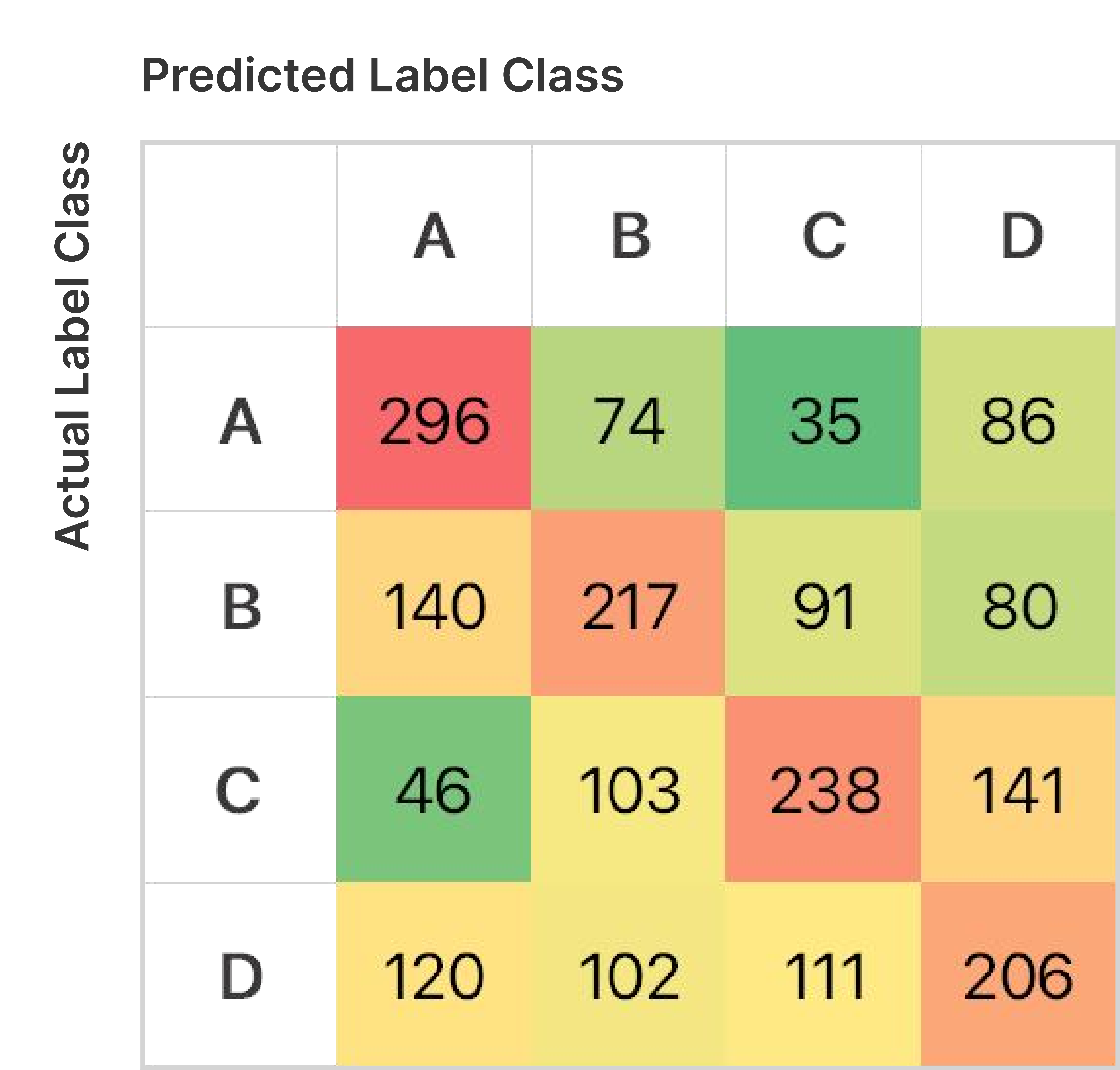}
    \caption{\textbf{RAPTOR with Collapsed Tree Querying}}
    \label{fig:confusionmatrixcollapsedtree}
  \end{subfigure}
  \caption{Confusion Matrices for RAPTOR with (a) HIRO Querying and (b) Collapsed Tree Querying on the QuALITY dataset}
  \Description{Confusion Matrices for RAPTOR with (a) HIRO Querying and (b) Collapsed Tree Querying on the QuALITY dataset}
  \label{fig:confusionmatrices}
\end{figure*}

\hyperref[fig:confusionmatrices]{Figure 3} presents the confusion matrices for RAPTOR with HIRO querying against Collapsed Tree querying on QuALITY dataset.

\textbf{Time Complexity Considerations:} \hyperref[tab:timecomplexity]{Table 4} details the time complexities for each hierarchical querying algorithm, highlighting the efficiency of HIRO querying with a linear time complexity \(O(n)\), compared to the logarithmic complexities of other methods. This efficiency is crucial for large-scale applications where processing speed is paramount.

For a more detailed analysis of space and time complexity, please refer to \hyperref[sec:complexityanalysis]{Appendix A}.

\section{Conclusion}

This research introduces Hierarchical Information Retrieval Optimization (HIRO), a novel approach that significantly enhances Retrieval-Augmented Generation (RAG) systems. By leveraging hierarchical document structures, HIRO optimizes context management for Large Language Models through recursive similarity thresholding and branch pruning, achieving a more efficient time complexity of O($n$). This method effectively balances context length, avoiding information overload while ensuring sufficient detail. HIRO demonstrates superior performance on the NarrativeQA dataset, surpassing traditional methods in key metrics. Although its performance on the QuALITY dataset shows slightly lower accuracy, it remains competitive. Notably, HIRO's optimization of context size results in substantial computational efficiency gains, reducing processing costs and paving the way for more sophisticated and resource-efficient AI applications.

\section{Reproducibility Statement}

\textbf{Language Model used for QA and Summarization:} OpenAI GPT 3.5 Turbo \cite{brown2020language} (\url{https://platform.openai.com/docs/models/gpt-3-5-turbo})\\

\noindent\textbf{Datasets Used:} NarrativeQA \cite{kocisky2018narrativeqa} (\url{https://github.com/google-deepmind/narrativeqa}), QuALITY \cite{pang2021quality} (\url{https://github.com/nyu-mll/quality})\\

\noindent\textbf{RAPTOR with HIRO Querying Mechanism (Source Code):} \url{https://github.com/krishgoel/hiro}\\

\noindent\textbf{RAPTOR (Original Source Code):} \url{https://github.com/parthsarthi03/raptor}

\bibliographystyle{ACM-Reference-Format}
\bibliography{main}

\appendix

\section{Complexity Analysis}
\label{sec:complexityanalysis}

In this section, we examine the computational and space complexities of three hierarchical retrieval algorithms—Tree Traversal Querying, Collapsed Tree Querying, and HIRO Hierarchical Querying.

\textbf{Tree Traversal Querying:} This method involves sorting nodes within each layer to select the top \(k\) nodes based on similarity. The sorting operation yields a time complexity of \(O(m \log m)\), where \(m\) is the number of nodes in the largest layer. Consequently, the overall time complexity is \(O(n \log m)\), with \(n\) representing the total number of nodes. The space complexity is \(O(n)\) due to the need to store the hierarchical structure.

\textbf{Collapsed Tree Querying:} This approach flattens the hierarchical structure into a list and sorts the nodes, leading to a time complexity of \(O(n \log n)\). The space complexity for storing the hierarchy is also \(O(n)\). While straightforward, this method may become less efficient with very large datasets.

\textbf{HIRO Hierarchical Querying:} HIRO employs a dynamic traversal strategy with adaptive thresholds, based on node similarity and parent context, achieving a linear time complexity of \(O(n)\) assuming constant times for similarity calculations and threshold adjustments. The space requirement includes \(O(n)\) for storing the hierarchy and \(O(n \cdot d)\) for node embeddings, where \(d\) is the dimensionality of the embeddings. Additional structures also require \(O(n)\) space, ensuring overall efficiency.

\begin{figure}
\label{fig:timecomplexity}
  \centering
  \includegraphics[width=\linewidth]{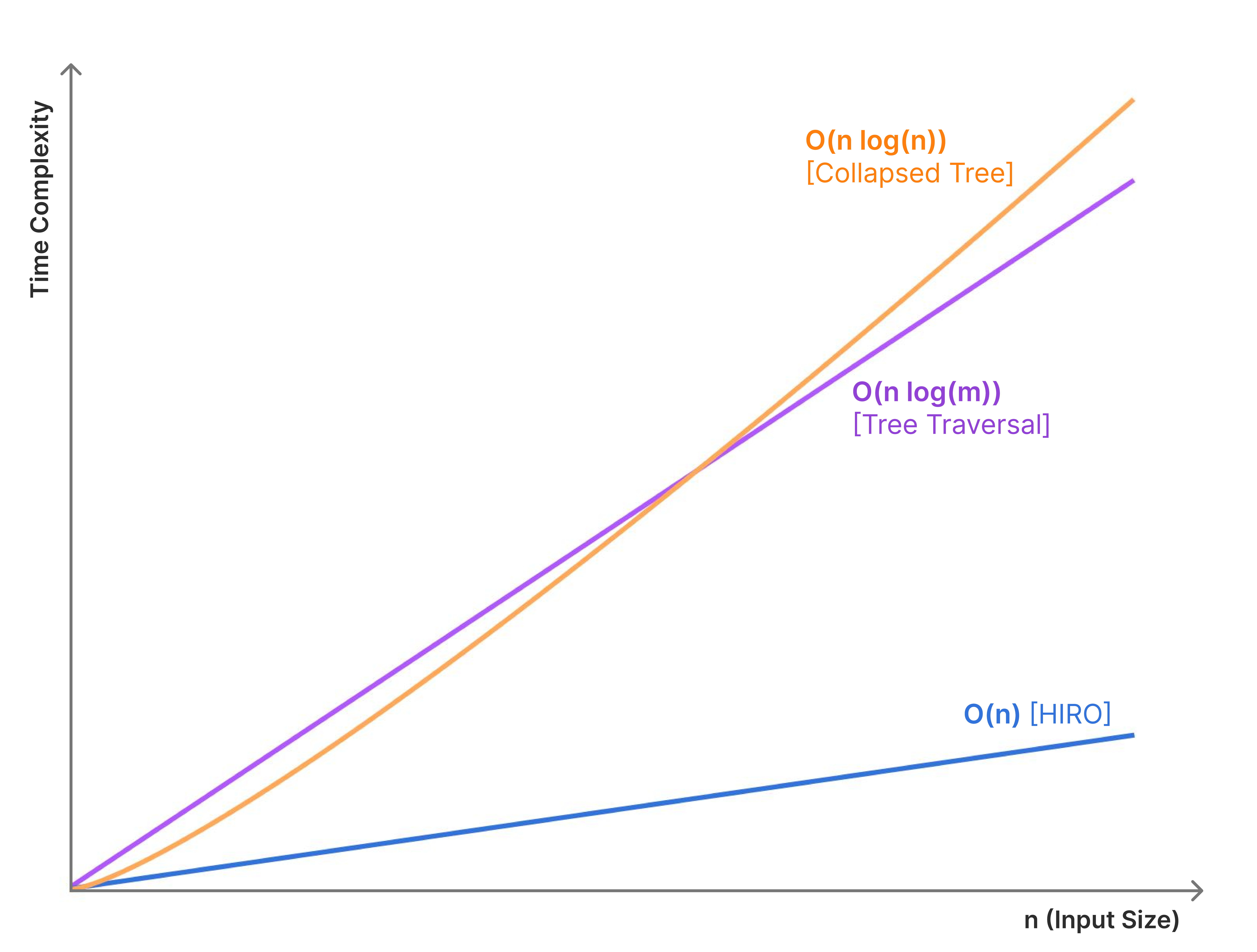}
  \caption{Comparison of Time Complexities for the Three Hierarchical Retrieval Algorithms.}
  \Description{Comparison of Time Complexities for the Three Hierarchical Retrieval Algorithms.}
\end{figure}

\hyperref[fig:timecomplexity]{Figure 4} illustrates the time complexities of these three hierarchical retrieval algorithms.





\begin{figure}
\label{fig:treetraversal}
  \centering
  \includegraphics[width=0.9\linewidth]{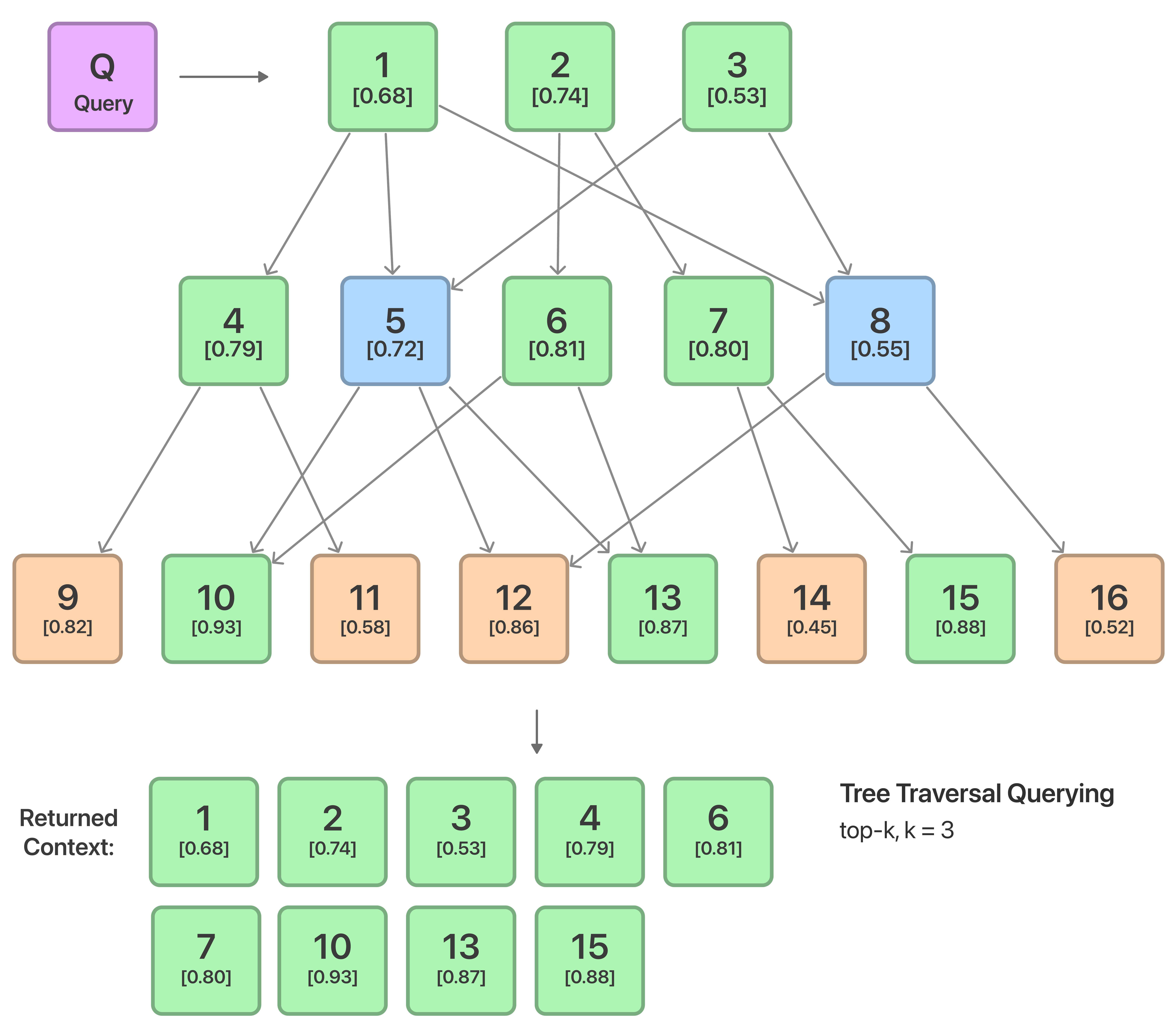}
  \caption{Tree Traversal Querying Mechanism. The top-$k$ root nodes are selected based on their similarity to the query. In the next layer, the children of these nodes are considered, and the top-$k$ nodes are again selected based on similarity. This process continues until reaching the leaf nodes.}
  \Description{The top-$k$ root nodes are selected based on their similarity to the query. In the next layer, the children of these nodes are considered, and the top-$k$ nodes are again selected based on similarity. This process continues until reaching the leaf nodes.}
\end{figure}

\begin{figure}
\label{fig:collapsedtree}
  \centering
  \includegraphics[width=0.9\linewidth]{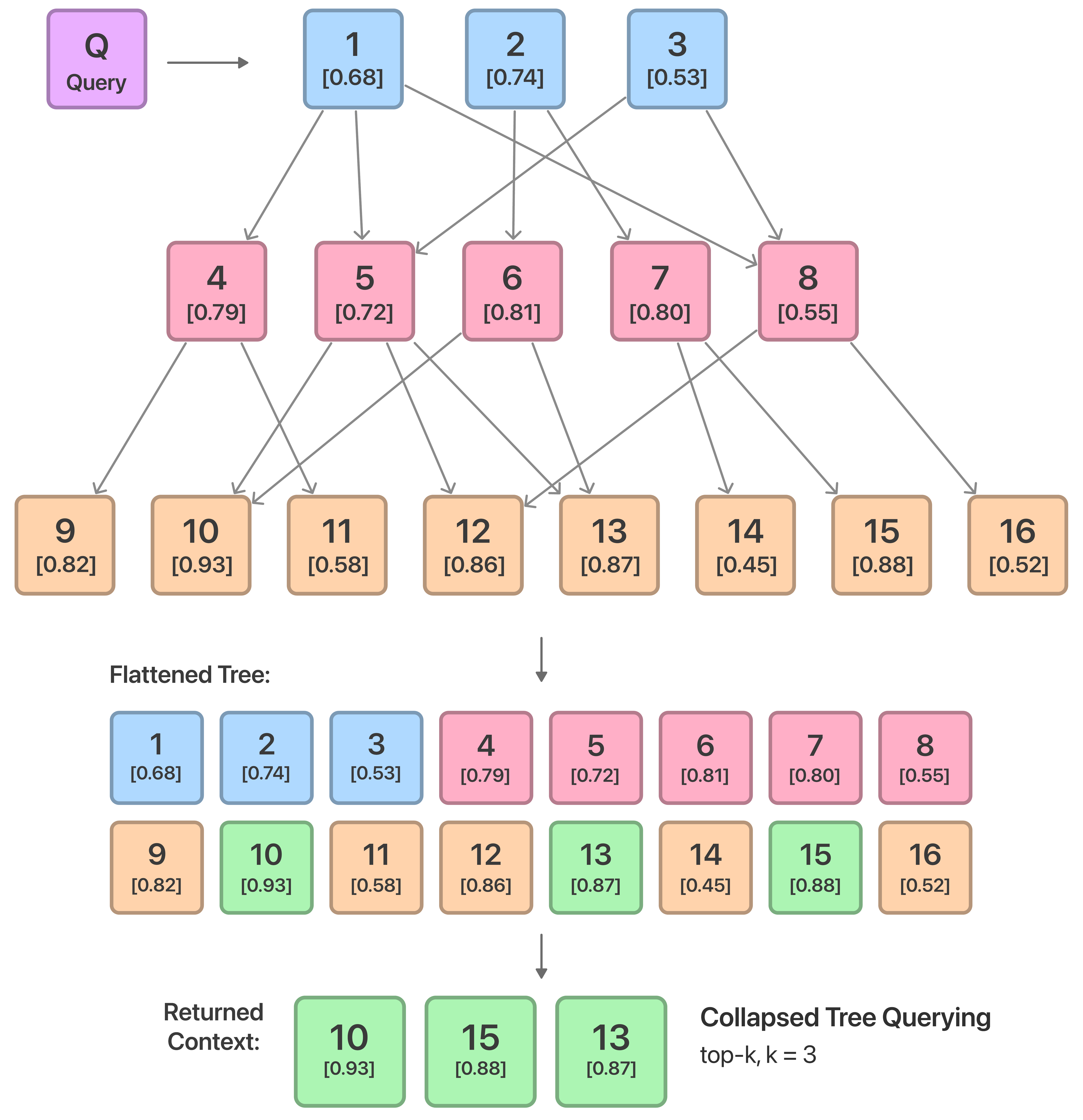}
  \caption{Collapsed Tree Querying Mechanism. The tree is flattened into a single layer, and the top-$k$ nodes are directly selected based on their similarity to the query embedding. The returned context consists of the most relevant nodes from this single-layer selection.}
  \Description{The tree is flattened into a single layer, and the top-$k$ nodes are directly selected based on their similarity to the query embedding. The returned context consists of the most relevant nodes from this single-layer selection.}
\end{figure}

\section{Production-Ready Pseudocode for HIRO Querying}
\label{sec:alternatepseudocode}
This revised version of the HIRO Querying pseudocode is designed for optimal integration into production environments. It replaces potentially complex while loops with more manageable for loops, ensuring both clarity and ease of maintenance. Importantly, it integrates two algorithms: HIRO Querying (\hyperref[alg:alternatehiroquerying]{Algorithm 3}) and its auxiliary Evaluate Children (\hyperref[alg:evaluatechildren]{Algorithm 4}), which is invoked within the HIRO algorithm. This configuration not only adheres to production standards but also facilitates easier debugging and scalability in operational systems.

While we utilize cosine similarity for node comparisons, other metrics like Euclidean distance, or Manhattan Distance can also be employed to measure vector distances.

\begin{algorithm}
\caption{\textbf{HIRO Querying}}
\label{alg:alternatehiroquerying}
\begin{algorithmic}
    \STATE\textbf{Input}: Query $query$, Tree $tree$, Selection Threshold $S$, Delta Threshold $\Delta$
    \STATE\textbf{Output}: Context Nodes
\end{algorithmic}
\begin{algorithmic}[1] 
\STATE $context \leftarrow []$
\STATE $earmarked \leftarrow []$
\STATE $l_{\text{current}} \leftarrow tree\text{.layer}[0]$
\FOR{node in $l_{\text{current}}$}
    \STATE $score \leftarrow \text{similarity}(query, node)$
    \IF{$score > S$}
        \STATE $earmarked.\text{append}(node)$
    \ENDIF
\ENDFOR
\STATE $context \leftarrow \text{evaluateChildren}(query, earmarked, S, \Delta)$
\STATE \textbf{return} $context$
\end{algorithmic}
\end{algorithm}

\begin{algorithm}
\caption{\textbf{Evaluate Children}}
\label{alg:evaluatechildren}
\begin{algorithmic}
    \STATE\textbf{Input}: Query $q$, Nodes $nodes$, Selection Threshold $S$, Delta Threshold $\Delta$
    \STATE\textbf{Output}: Local Context Nodes
\end{algorithmic}
\begin{algorithmic}[1] 
\STATE $local\_context \leftarrow []$
\FOR{parent\_node in $nodes$}
    \STATE $parent\_similarity \leftarrow \text{similarity}(q, parent\_node)$
    \FOR{node in $parent\_node.\text{children}$}
        \STATE $score \leftarrow \text{similarity}(q, node)$
        \STATE $delta \leftarrow score - parent\_similarity$
        \IF{($delta > \Delta$) \OR ($node\text{.is\_leaf()}$ \AND $score > S$)}
            \STATE $local\_context.\text{append}(node.\text{content})$
        \ELSE
            \STATE $local\_context.\text{extend}(\text{evaluateChildren}(q, [node], S, \Delta))$
        \ENDIF
    \ENDFOR
\ENDFOR
\STATE \textbf{return} $local\_context$
\end{algorithmic}
\end{algorithm}

\section{Bayesian Optimization for Fine-Tuning of Hyperparameters}
\label{sec:bayesianoptimisation}

In our efforts to optimize the performance of our model on various datasets, we focus on fine-tuning key hyperparameters. The target function for optimization is defined as the performance of the model, which we quantify through a weighted or normalized sum of various metrics relevant to each specific dataset. These metrics may include BLEU Scores, ROUGE Scores, METEOR Scores, Precision, Recall, F1 score, or other dataset-specific performance indicators. The performance function being optimized for each dataset is provided below.

We utilize Bayesian Optimization to fine-tune the hyperparameters Selection Threshold ($S$) and Delta Threshold ($\Delta$) for various datasets. Bayesian Optimization is chosen for its proficiency in optimizing complex, multi-metric objective functions \cite{snoek2012practicalbayesianoptimizationmachine}. Bayesian Optimization utilizes a probabilistic model to estimate the objective function landscape and an acquisition function to efficiently explore this landscape by focusing on areas likely to offer improvements.\\

\textbf{NarrativeQA Dataset Performance Function:}The performance function for the NarrativeQA dataset is designed to evaluate the model's ability to comprehend and summarize narratives. This function incorporates the following key metrics, ensuring thorough evaluation across different dimensions of summarization quality. To maintain consistency and comparability, each metric score is normalized individually.
\begin{itemize}
    \item \textbf{ROUGE (Recall-Oriented Understudy for Gisting Evaluation):} Measures the overlap of n-grams between the generated summaries and reference summaries to assess the recall of content.
    \item \textbf{BLEU-1 and BLEU-4 (Bilingual Evaluation Understudy):} BLEU-1 measures the match of single words (1-gram) between the generated and reference texts, focusing on lexical accuracy. BLEU-4 extends this evaluation to 4-grams, assessing the coherence and order of longer phrases in the generated text.
    \item \textbf{METEOR (Metric for Evaluation of Translation with Explicit ORdering):} Evaluates translation QuALITY by aligning the generated text with reference texts, accounting for synonymy and stemming, and applying varied weights to different matching components.
\end{itemize}

\[P = ROUGE + {BLEU}_{1} + {BLEU}_{4} + METEOR\]\\

Normalization process ensures that each metric contributes equally to the performance function, allowing for a balanced and fair evaluation of the model's capabilities.\\

\textbf{QuALITY Dataset Performance Function:} For the QuALITY dataset, the performance function focuses exclusively on the F1 score to assess the model's capability in understanding and accurately answering multiple-choice reading comprehension questions. The F1 score is a balanced metric that considers both precision and recall, making it an ideal choice for evaluating the model's overall effectiveness in this context.\\

\[P = F1\]
\[\implies P = 2 \cdot \frac{Precision \cdot Recall}{Precision + Recall}\]

\end{document}